\documentclass[10pt,twocolumn,letterpaper]{article}

\usepackage{cvpr}
\usepackage{times}
\usepackage{epsfig}
\usepackage{graphicx}
\usepackage{amsmath}
\usepackage{amssymb}
\usepackage{booktabs}
\usepackage{comment}
\usepackage{dsfont}
\usepackage{microtype}
\usepackage[ruled,linesnumbered,lined,commentsnumbered]{algorithm2e}
\usepackage[subrefformat=parens]{subcaption}

\usepackage[pagebackref=true,breaklinks=true,letterpaper=true,colorlinks,bookmarks=false]{hyperref}

\makeatletter
\newcommand{\printfnsymbol}[1]{%
  \textsuperscript{\@fnsymbol{#1}}%
}

\makeatother

\cvprfinalcopy 
\def\cvprPaperID{8114} 

\begin{document}
\title{Contrastive Learning with Large Memory Bank and Negative Embedding Subtraction for Accurate Copy Detection}
\author{
Shuhei Yokoo\\
DeNA Co., Ltd.\\
{\tt\small shuhei.yokoo@dena.com}
}

\maketitle

\begin{abstract}
Copy detection, which is a task to determine whether an image is a modified copy of any image in a database, is an unsolved problem.
Thus, we addressed copy detection by training convolutional neural networks (CNNs) with contrastive learning. Training with a large memory-bank and hard data augmentation enables the CNNs to obtain more discriminative representation. Our proposed negative embedding subtraction further boosts the copy detection accuracy.
Using our methods, we achieved 1st place in the Facebook AI Image Similarity Challenge: Descriptor Track. Our code is publicly available here: \url{https://github.com/lyakaap/ISC21-Descriptor-Track-1st}
\end{abstract}

\begin{figure}[t]
\centering
\vspace{0.5mm}
  \includegraphics[width=\linewidth]{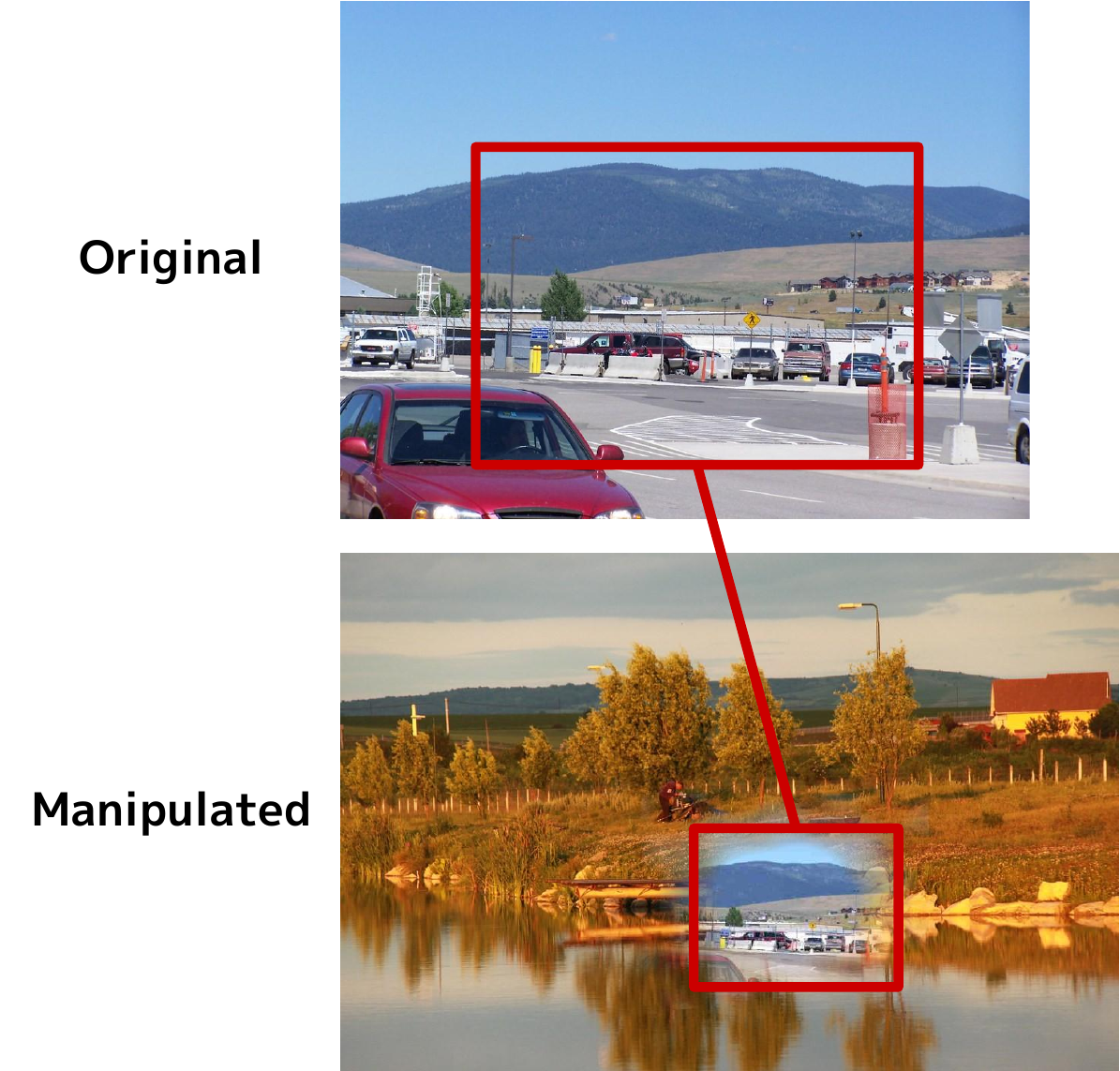}
  \vspace{0.5mm}
  \caption{
  Example of manipulation from the dataset for the Image Similarity Challenge 2021.
  The original image (top) is overlaid in the small part of the manipulated image (bottom), which hinders its identification in a large database.
  Our model successfully detected the copy and identified the original image in this example.
  Credit of images (user name of Flickr): CrusinOn2Wheels (original image), bortescristian (background image used for manipulation).
  }
  \label{fig:teaser}
\end{figure}

\section{Introduction}
In recent years, with the development of social media, the problems of plagiarism and unauthorized copy have become more serious. Accordingly, attention to research on copy detection, which is a task to determine whether an image is a modified copy of any image in a database, has increased~\cite{ISC2021,Jiang2014VCDBALCopy,Christlein2012AnEOCopy}.

In this study, we propose a strong copy detection pipeline that consists of EfficientNetV2~\cite{EfficientNetV2} trained with contrastive learning and a post-process that effectively utilizes negative embeddings.
Our training pipeline has multiple steps inspired by progressive learning~\cite{EfficientNetV2}, which is a model training technique that increases input image resolution and regularization as a training step proceeds.
In each step, models are trained by contrastive learning with hard data augmentation to obtain discriminative representation.
Embeddings extracted by our model are processed employing our proposed negative embedding subtraction, which improves the copy detection performance by isolating a target sample from similar negative samples.
Using our methods, accurate copy detection is realized even in difficult samples as shown in Figure \ref{fig:teaser}.

Our contribution is summarized as follows: (1) multi-step training with contrastive learning and large memory-bank, (2) carefully designed data augmentation strategy that precisely reproduces manipulated images from the dataset, (3) novel post-process method that enhances embeddings utilizing negative samples, (4) results that significantly outperform baseline methods on a challenging dataset, enabling to win the Facebook AI Image Similarity Challenge: Descriptor Track at NeurIPS’21.

\begin{figure*}[t]
\centering
  \includegraphics[width=\linewidth]{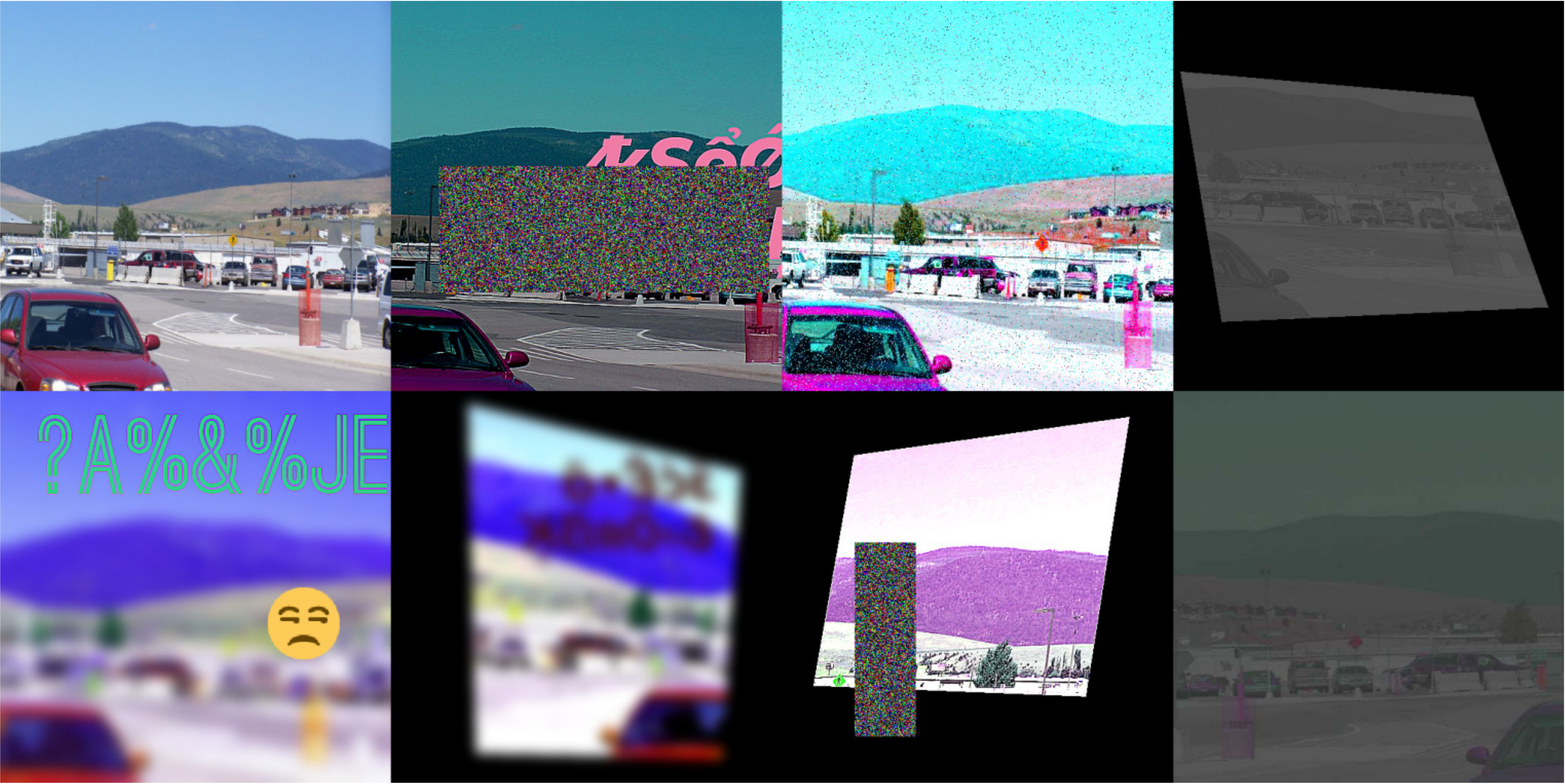}
  \caption{
    Examples of images processed by our data augmentation pipeline.
    The image at the top left is the original image, and the others are processed images.
    An exhaustive list of data augmentation is shown in Section~\ref{sec:aug}.
  }
  \label{fig:aug}
\end{figure*}

\section{Dataset}
During the Facebook AI Image Similarity Challenge (ISC21), a new dataset for copy detection, \textit{DISC21}~\cite{ISC2021} was released.
The DISC21 is composed of 1 million reference images, 1 million training images, and 50,000 query images.
A subset of the query images was derived from the reference images, and the rest of the query images were not.
Both the query and reference set contain a majority of ``distractor'' images that do not match.
The goal of the ISC21 was to distinguish ``distractor'' images and identify which reference images are used for the query images.

Ground truth pairs between query images and corresponding reference images were also provided.
However, augmenting the query and reference images for training was prohibited by the competition rules.
Therefore, participants of the competition were encouraged to use the training images for model training, as augmenting the training images was permitted for any usages.

\section{Method}

\subsection{Data Augmentation}
\label{sec:aug}

Our augmentation pipeline consists of crop, rotation, horizontal flip, vertical flip,  padding, aspect ratio change, perspective transform, overlay onto background image, overlay of text and emoji, changes in brightness, saturation and grayscale, random erasing~\cite{RandomErasing}, blur, color palette with dithering, JPEG encoding, edge enhance, pixelization, and pixel shuffling.
The order to apply augmentation is continuously and, randomly shuffled to generate more diversified examples.
These augmentations were selected based on how the dataset for ISC21 was created.
Parameters of the augmentations were manually set to the range, as close as possible to the actual copied examples by eye-checking.
Examples of images processed by our data augmentation pipeline are shown in Figure~\ref{fig:aug}.

As observed, the processed examples were drastically changed from the original image and barely recognizable.
The final evaluation set (query set of phase-2) contained more difficult examples than the evaluation set provided during the competition period, thus we believe such drastic data augmentations contributed to the final result.

\begin{table*}[t]
\centering
\begin{tabular}{ccccccc}
\toprule
Input Resolution & Augmentation & Trained w/ Reference & Trained w/ GT & Post-process & \textmu AP & Recall@P90 \\
\midrule
256$\times$256 & weak & & & & 0.5831 & 0.4644 \\
384$\times$384 & intermediate & & & & 0.6231 & 0.5237 \\
512$\times$512 & strong & \checkmark & & & 0.6435 & 0.5662 \\
512$\times$512 &  & \checkmark & \checkmark & & 0.7557 & 0.6404 \\
512$\times$512 &  & \checkmark & \checkmark & \checkmark & \textbf{0.7743} & \textbf{0.6892} \\
\bottomrule
\end{tabular}
\vspace{0.2mm}
\caption{
    Comparison of each step, including post-processing step.
    We report \textmu AP (micro-average precision) and Recall@P90 (Recall at Precision 90) in the private-set of phase-1.
    Higher values of these metrics are better.
    ``Augmentation'' means the magnitude of the data augmentation. If the column of ``Augmentation'' is empty, it means no data augmentation was applied.
    ``Trained w/ Reference'' means using reference images for negative pairs of contrastive learning, and ``Trained w/ GT'' means using ground-truth pairs of the  public set of phase-1.
    ``Post-process'' is our proposed negative embedding subtraction.
    The results shown here are based on the previous row (except for the first row).
}
\label{tab:ablation}
\end{table*}

\subsection{Training}

For model training, we employ contrastive loss~\cite{ChopraHL05ContrastiveLoss} with cross-batch memory~\cite{Wang2020CrossBatchMF}.
Our initial experiments showed this combination of methods outperforms other metric learning losses, such as triplet loss~\cite{FaceNetTripletLoss15,HofferA14TripletLoss2,WangSLRWPCW14TripletLoss1} and AP loss~\cite{LandmarkListWiseLoss}.
By using the cross-batch memory, our model can train with more beneficial and diversified negative pairs.

Positive pairs are created from a single training image, as in the self-supervised learning methods~\cite{He2020MomentumCF,simsiam}.
Specifically, the first sample is generated by applying our data augmentation pipeline described in Section~\ref{sec:aug}, and another one is generated by applying standard data augmentations including resize, crop, and flip.
For the negative pairs, all possible combination pairs are used except the positive pairs.

The model training is conducted in multiple steps inspired by progressive learning~\cite{EfficientNetV2}.
As the training steps proceed, the input resolution and the magnitude of the data augmentation are increased.
We increase the magnitude of the data augmentation by expanding the range of transformation and increasing the probability of application of transformations.

The entire training procedure is provided as follows.
In the first step, we start training with a small input resolution and weak data augmentation.
In the next step, the model trained in the first step is fine-tuned, where the input resolution and the magnitude of the data augmentation are increased.
Next, we use both the training images and the reference images as a negative pair for training.
Finally, the positive pair formed by the query images and the reference images is also used for training.
In the last two steps, we did not perform data augmentation on the reference images and the query images, because this is prohibited by the rules.
For more information regarding each step, refer to Table~\ref{tab:ablation}.

\subsection{Post-process}
In the Descriptor Track of ISC21, we were required to directly submit image descriptor.
Thus, some optimization approaches such as similarity normalization~\cite{Jgou2011ExploitingDD} is not applicable.
Moreover, interaction between other query images are prohibited by the rules of the competition, well-known re-ranking methods such as query expansion and diffusion \cite{ChumPSIZ07AQE,ArandjelovicZ12DQE,ToliasJ14HQE,DonoserB13Diffusion} are also unavailable.

However, the use of the training set is allowed.
The training set was designed as a statistical twin of the reference set, and was carefully collected such that there were no overlaps between the training set and the reference set.
This means samples from the training set are always negative samples (not copied) for the query set, and samples from the training set that are similar to a query sample can be regarded as hard negative examples.
Exploiting this nature, we propose a post-process method in which the query sample is isolated from such examples by vector subtraction, named as \textit{negative embedding subtraction}.

Our negative embedding subtraction is simple, but effective for obtaining discriminative representation in feature space. Algorithm~\ref{algo} illustrates the procedure of our proposed method, which has three hyperparameters: number of iterations~$n$, top-k of $k$-NN search $k$, and subtraction factor~$\beta$. We set $n$ to 1, $k$ to 10, and $\beta$ to 0.35 respectively.

\begin{algorithm}[htbp]
\caption{Negative embedding subtraction.}
\label{algo}
\DontPrintSemicolon
\SetAlgoLined
\KwIn{
         target image descriptor~$x$,
         negative image descriptor set~$\mathcal{X}_\mathrm{neg}$,
         number of iterations~$n$,
         top-k of $k$-NN search $k$,
         subtraction factor~$\beta$
}
\KwOut{processed image descriptor~$x$}
\BlankLine
\For{$i = 0~\mathrm{to}~n - 1$}{
   search $\mathrm{NN}_{k}(x) \subset \mathcal{X}_\mathrm{neg}$ \;
   \For{$x_{neg} \in \mathrm{NN}_{k}(x)$}{
      $x \leftarrow x - \frac{\beta}{k} x_{neg}$ \;
   }
   $x \leftarrow x / \|x\|_2$ \;
}
\Return $x$ \;
\end{algorithm}

\begin{table}[t]
\centering
\begin{tabular}{l c c}
\toprule
Team & \textmu AP & Recall@P90 \\
\midrule
titanshield2             & 0.7418  & 0.7018    \\
\textbf{lyakaap (Ours)}         & \textbf{0.6354}  & \textbf{0.5536}   \\
S-square & 0.5905  & 0.5086    \\
visionForce & 0.5788 & 0.4886 \\
forthedream2 & 0.5736 & 0.4980 \\
\bottomrule
\end{tabular}
\vspace{0.2mm}
\caption{Leaderboard with final results (only top 5 teams are listed here).
Our results are in bold.}
\label{tab:lb}
\vspace{0.5mm}
\end{table}

\section{Experiments and Results}

\subsection{Implementation Details}
Details for model architecture and training settings basically follow~\cite{GLR20191stplace,TwoStageDiscriminativeReranking}.
We employ ``tf\_efficientnetv2\_m\_in21ft1k'' in timm~\cite{rw2019timm} library as a backbone.
The weights of the backbone were pre-trained on the ImageNet-21K~\cite{imagenet}.
Model training is conducted by using the stochastic gradient
descent with momentum set to 0.9.
The dimension of the descriptor is set to 256, which is specified by the competition rules.
For contrastive loss, we set 0.0 for the positive margin, 1.0 for the negative margin, and 20,000 for the memory size.
For data augmentation, we use AugLy~\cite{bitton2021augly} library.
For nearest neighbor search, we use faiss~\cite{JDH17} library.

\subsection{Comparison of Each Step}
We evaluate the performance of each training step and post-process using the private-set of phase-1.
Table~\ref{tab:ablation} shows the evaluation results of each-step.
All steps considerably improves the performance, particularly the step ``Trained w/ GT''.
Training with the ground truth pairs may be highly beneficial, as these pairs contain images manually edited, which is difficult to replicate with automated data augmentation.
Our proposed post-process further improves the performance by a large margin, in spite of its simplicity.

\subsection{Competition Results}

The final results of the competition are shown in Table~\ref{tab:lb}.
As the team of the top score did not share their entire solution, we ranked 1st place among the participants holding the prize eligibility.

In addition, we experiment replacing the negative reference images with the training images in the last two training steps. This provides a score of 0.6297 with our post-process, only degrading 0.0057 from the original score. This suggests that our copy detection solution does not overfit to the reference set.

\section{Conclusion}
We presented a strong pipeline for copy detection based on contrastive learning with a carefully designed data augmentation pipeline.
Our work leverages recent approaches and we proposed a novel post-process method that shows significant improvements. 
Using our methods, we achieved 1st place in the Facebook AI Image Similarity Challenge: Descriptor Track.


\clearpage
{\small
\bibliographystyle{ieee_fullname}
\bibliography{egbib}
}

\end{document}